\documentclass[twocolumn]{article}
\usepackage{microtype}
\usepackage{times}
\usepackage{latexsym}
\usepackage{graphicx}
\usepackage{amsmath}
\usepackage{url}
\usepackage{subcaption}
\usepackage[hang,flushmargin]{footmisc}  
\usepackage[ruled,vlined,linesnumbered]{algorithm2e}

\title{\textbf{Ten-year Survival Prediction \\ for Breast Cancer Patients}}
\author{
       Changmao Li, Han He, Yunze Hao, Caleb Ziems \\
       Emory University\\
       \textit{\{changmao.li, han.he, yunze.hao, cziems\}@emory.edu}
}
\date{}
\begin{document}
\maketitle

\section{Introduction}
Different stages of breast cancer require different treatments. Understanding the current stage of a patient's breast cancer is crucial then for applying the best treatment. For this purpose, machine learning models may be used to learn and predict patient survival or other clinical outcomes from professionally-labeled features or large-scale genomic profiles \cite{cheng2013development, yousefi2017predicting}.

In the present study, we train and tune models to predict the 10-year survival of breast cancer patients using the METABRIC (Molecular Taxonomy of Breast Cancer International Consortium) dataset. This dataset includes both hand-labeled clinical data and high-dimensional genomic data from over 2,000 patients. We trained our first set of models on the clinical data and our second set of models on genomic data. Our goal was to construct a model from the latter set which could outperform our best model from the former. 

We defined our learning objective as a classification problem with binary class labels. \textbf{Class 1} was assigned to patients who had died of the disease within 10 years (120 months) of prognosis. \textbf{Class 2} was assigned to patients who survived longer than 10 years.

The nature of the METABRIC data makes our learning problem a challenge. We recognize that the clinical features were hand-selected by experts and refined through decades of medical research. Thus, we might reasonably expect many of these features to be strong predictors of patient outcomes. Genomic data, on the other hand, is quite noisy, and may lead our models to overfit because the number of genomic features far outweighs the number of samples. For this reason, we implemented four advanced learning algorithms which we selected to overcome this challenge. These include semi-supervised learning, L1-regularized logistic regression, a multi-layer perceptron with grasshopper optimization, and XGboost. We also trained a number of baseline models. By comparing these with our advanced algorithms, we had a better understanding of our problem, which contributed to the development of some best-approaches for reaching our goal. All our implementations are public available. \footnote{\url{https://github.com/FrankLicm/breast-cancer-prediction}} 




\section{Exploratory Data Analysis}

Our analysis of the METABRIC data was used to guide our preprocessing steps. Most notably, we find that the data contains a number of irregularities, missing values, and unnecessary features which needed to be addressed. To avoid biasing our data through improper deletions or imputations, we need to show that the data is Missing at Random (MAR) or Missing Completely at Random (MCAR). This is especially important in the case of clinical data, which contains a large number of missing values.

\subsection{Clinical Patient Data}

The clinical data contains entries for 2,509 patients. Each entry contains 17 explanatory variables and three response variables. We see that 14 of the 17 explanatory variables come from discrete categories including the genomic classification of the cancer (ER/PR/HER2), the estrogen receptor status, and a binary indication of chemotherapy treatment. In our preprocessing steps, we will encode these categories using dummy variables. 

The remaining three explanatory variables contain continuous numerical data from the Nottingham prognostic index (NPI), the patient’s age at diagnosis, and a count of the observed cancer-positive lymph nodes. Since these features will be used in scale-sensitive models like logistic regression, we will need to standardize them first. 

The three response variables provide the overall survival time or the time a patient was last seen alive, their status (either living or dead), and whether breast cancer was the cause of their death. These variables can be used to determine our binary labels. We label patients \textbf{Class 1} if they died within $t \leq 120$ months and the cause of their death was breast cancer. We label patients \textbf{Class 2} if they were last seen $t>120$ months after prognosis. For those who died of other causes or were still living and were last seen within $t < 120$ months, we did not assign them a label. We will consider them later on, however, in our implementation of semi-supervised learning. 

Now we turn to missing values. Each column in the clinical patient data contains anywhere from 11 to 745 missing values, with a median number of 529. Our first task is to decide whether these are missing at random. If the MAR condition is satisfied, we can safely drop the missing values without biasing the data.

Our first indication that MAR is satisfied comes from the correlation matrix in Figure 1. Originally, this plot contained some correlations of +1.00. However, the variables were only one missing value. We decided that there was not sufficient evidence that their nullity was correlated, so we dropped them from the plot. Now, all but lymph node / histological subtype show very low correlations. Since histological subtype is a categorical value, we will avoid any bias from deletion by simply adding a new "null" category to the dummy variable. This strategy can be extended to the other categorical variables as well. For the numeric lymph node variable, we will have to consider other options. 

Since the lymph node count and the histological subtype are both missing in many common instances, we cannot determine whether the nullity of one predicts the outcome of the other. Instead, we consider the effect of null values on the outcomes of our continuous variables. In Figure 2, we use box plots to show that the continuous values of NPI and age are not correlated with the nullity of the lymph node count. Although it is impossible to prove MAR with certainty, these plots provide adequate evidence for the MAR condition. It is safe to drop the missing values for lymph node count. We have determined that deletion is preferable to imputation in this case since the distribution of lymph node counts is not symmetric, but instead skewed towards zero.

\begin{figure}
\centering
  \includegraphics[width=\linewidth]{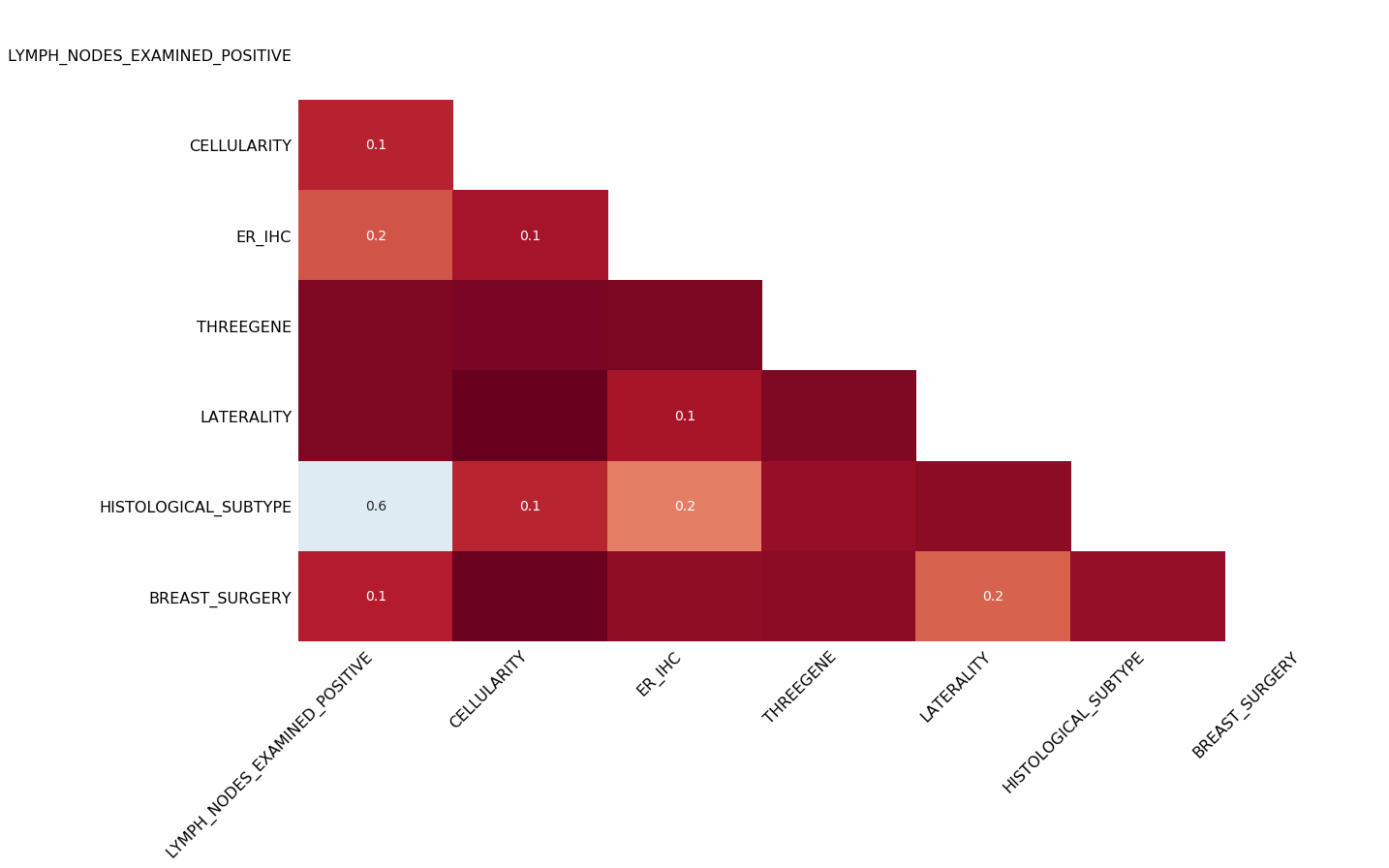}
\caption{\textbf{Missing values correlation matrix for clinical data. Low correlations are dark and high correlations are light. We observe low correlations between all features except Lymph Node Count and Histological Subtype.}}
\label{fig:test}
\end{figure}

\begin{figure}
\centering
\begin{subfigure}{0.25\textwidth}
  \centering
  \includegraphics[width=\linewidth]{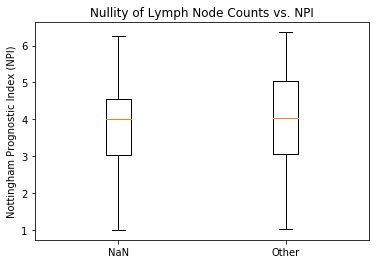}
  \caption{NPI}
  \label{fig:sub1}
\end{subfigure}%
\begin{subfigure}{0.25\textwidth}
  \centering
  \includegraphics[width=\linewidth]{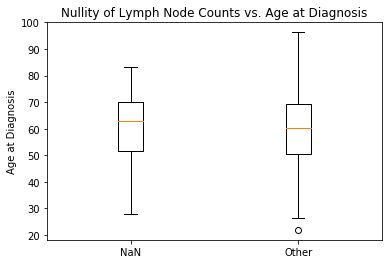}
  \caption{Age}
  \label{fig:sub2}
\end{subfigure}
\caption{\textbf{Nullity of lymph nodes examined positive vs. NPI and age at diagnosis.}}
\label{fig:test}
\end{figure}

\subsection{Gene Expression Data}
The METABRIC dataset provides gene expression data for 1,903 patients. Gene expression is given as the abundance of mRNA for that gene, and we have data on 24,367 genes. We see that the number of features is much greater than the number of observations. Therefore high variance and overfitting are both concerns. We will rely on the L1 regularization for automatic feature selection in the regression models. When the regularization coefficient is large enough, many coordinate weights will be zero. We will also consider XGBoost, which handles overdetermined learning problems well. And to deal with missing values, we can again remove them since there are only 11 patients with missing values.

\subsection{Mutations Data}
Gene mutations can lead to significant consequence including null mutations, abnormal protein product, etc. Germline mutations in BRCA1/BRCA2, are known to considerably increasing the risk of occurring breast cancer and ovarian cancer in familial cases \cite{lux2006hereditary}. Although the mutations data contains a large number of features, we only consider the coding region of the mutation (Hugo symbol) and the variant of the classification (missense mutation, nonsense, nonstop, etc.)

\subsection{CNA Data}
The CNA data gives us the number of copies for 22,543 genes across 2,174 patients. A copy number of -2 indicates both copies have been deleted, and -1 means one has been deleted, and 0 means the patient has both copies. However, a 1 indicates a patient with more than two copies and a 2 indicates one many more than two copies. For this reason, we cannot interpret these values as linearly ordered. Instead, they can be treated as categories using a one-hot encoding scheme.

\section{Preprocessing Procedure}
One of our purposes is to compare the performances on clinical data and genomic data, so how we preprocess them is key to our comparison equality. For this purpose, we keep the same patients between the genomic data and clinical data. 

For preprocessing clinical data, first we need to removed the 3 response variables in the clinical which are last seen alive, status (living/dead) and cause of their death. For dealing with the missing values, because we don't want to fake data, we drop the missing value rows of numeric value and drop the missing value row whose number of missing value is larger than 2. For those non-deleted missing categorical values, we set the null value as a new category of those categorical values.
After that, we encode the categorical value with one-hot encoding methods and standardize features by removing the mean and scaling to unit variance. The standard score of a sample x is calculated as:
\begin{equation}
        z = (x - u) / s
\end{equation}
where u is the mean of the training samples or zero,  and s is the standard deviation of the training samples.

For preprocessing genomic data, first we extract the patient in both clinical data and genomic data. For each patient, we try to build input matrix from all three genomic tables. For Gene expression and CNA, since they are all numerical value, we directly add them as new columns. For DNA mutations, we only use variant classification which is a categorical value and encode them using one-hot encoding. We delete all missing value columns since we do not want to fake data. After that we apply feature scaling as the clinical data. After preprocessing, we have 1,393 patients, 

\section{Validation Procedure}

Once our data was processed, cleaned, and normalized, we randomly shuffled the data and split samples into 80\%, 10\%, and 10\% for training, validation, and testing respectively. The same split was used for all experiments. Where time permitted, we utilized the full 5-fold nested cross validation procedure outlined in the following subsection. This procedure was implemented for all baseline models as well as the L1 Logistic Regression and XGBoost. Due to time constraints, we were not able to run the same validation procedure for the MLP classifier with Grasshopper Optimization. In this case, we ran experiments on the validation and testing sets three times and computed average scores to get a fair evaluation.


\subsection{Nested Cross Validation}

\begin{figure*}[ht!]
\centering
  \includegraphics[width=\linewidth]{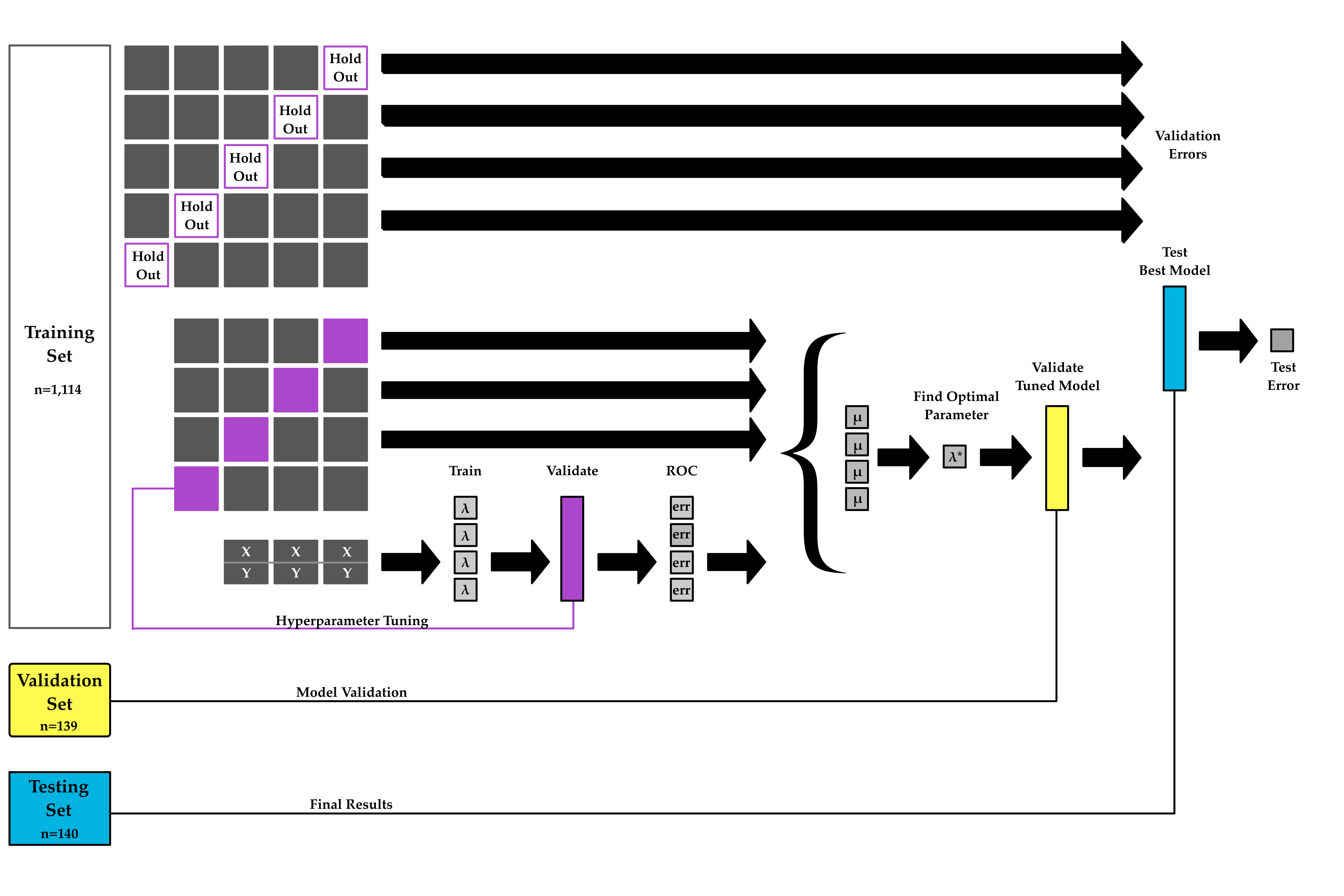}
\caption{\textbf{5-Fold Nested Cross Validation Procedure}}
\label{fig:test}
\end{figure*}

Here, we give a detailed outline of our 5-fold nested cross validation procedure. In Figure 3, we provide a useful schematic diagram to help understand the flow of data.

Our 80-10-10 split gives us 1,114 training samples, 139 validation samples, and 140 testing samples, and the training set was split into 5 folds. At each iteration of the outer loop, one of the 5 folds was held out. The remaining 4 folds were used in an inner loop for computing our accuracy curves. At each iteration of the inner loop, one of the 4 folds was reserved for validating models. The other three were used to train the models along a logarithmic or linearly spaced array of hyperparameters. This produced accuracy curves with a mean accuracy and standard deviation given for each parameter choice. We chose our optimal parameter such that it provided the simplest model and had an accuracy bounded within one standard deviation of the best mean accuracy.

After running through the inner loop, we validated the tuned model using the validation dataset. Upon completion of the outer loop, we were left with 5 optimal parameter choices and 5 validation errors. For validation scores, we report the average of these 5 scores. From the optimal parameters, we chose the parameter with the best validation error and used this to train a best model. The best model was trained using all of the data from both the training and validation sets. We report the last score using our testing dataset.

\subsection{Evaluation metrics}

Our original data was somewhat imbalanced, with 735 / 1,125 (or 65\%) of the labels belonging to Class 2. To address this class imbalance, we will not use mean squared errors for validation and testing, but instead, we will rely on the Receiver Operator Characteristic (ROC).

These curves plots two parameters: True Positive Rate and False Positive Rate. An ROC curve plots TPR vs. FPR at different classification thresholds. Lowering the classification threshold classifies more items as positive, thus increasing both False Positives and True Positives. We measure the entire two-dimensional area underneath the ROC curve as our final report score of each algorithm. These scores are given the name AUC for "area under the curve." An AUC of 0.5 is considered "random guessing" and an AUC of 1.0 describes a perfect model.

\section{Baseline Algorithms}

We chose our baseline models to represent a wide range of machine learning approaches, from the non-parametric K-nearest neighbors algorithm and ensemble methods to generalized linear models and a neural architecture. For each model, we provide insights on performance and compare results across models.

\subsection{K-Nearest Neighbors}
Our first baseline model was the simple K-nearest neighbors algorithm. This algorithm takes the $k$ data points from the training set which are closest to the unknown type. It then aligns its decision for that type with the best represented class among its neighbors. 

This approach was not expected to perform well for either dataset. The clinical data is primarily categorical, so a large number of datapoints lie along the edges and corners of the feature space. This reduced the significance of the euclidian distance calculation in K-NN. 

For the genomic data, we anticipated what is known as the "curse of dimensionality." Although features were not categorical, the high dimension of the genomic data produced a similar problem of sparsity. The volume of the feature space was so large that every neighboring data point would be far removed from the unknown type. As a result, no meaningful notion of similarity could be established. 

Although we tuned the $k$ parameter, we did not see any success. The average AUC validation score for the clinical data was 59.6, and the test score was 55.8. The average validation score for the genomic data was 50.6 and the test score was 52.1. These are given in Tables 1 and 2. 

Low AUC scores confirm two major problems in the K-NN approach. We see that, for the genomic data, the model did not perform significantly better than random chance. Though increasing $k$ reduced the observed variance between validation folds, there not enough information to learn the data with this approach, so the model was underfit.

\subsection{Logistic Regression}

Next, we implemented a logistic regression classifier with stochastic gradient descent and an elastic net mixing parameter of $\alpha = 0.95$. The heavily weighted L1 penalty allows the model to achieve a sparse solution, but the additional L2 penalty helps reduce irregularities from large model coefficients. We did not choose a purely L2 regularized model because it would be more vulnerable to outliers and would likely overfit the genomic data, using many non-zero coefficients.

Our cross-validation procedure was used to tune the regularization coefficient $\lambda$. In Figure 4, we see that a very small $\lambda$ results in low bias but high variance. This indicates overfitting. We aimed to increase $\lambda$ to reduce the complexity of the model. When $\lambda$ became too large, all of the model weights were forced to zero. In this case, the model made a constant class prediction and achieved an AUC of 50.

\begin{figure} [b!]
\centering
\begin{subfigure}{0.25\textwidth}
  \centering
  \includegraphics[width=\linewidth]{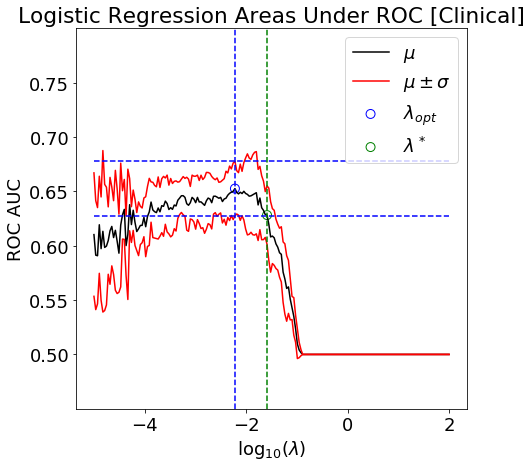}
  \caption{clinical}
  \label{fig:sub1}
\end{subfigure}%
\begin{subfigure}{0.25\textwidth}
  \centering
  \includegraphics[width=\linewidth]{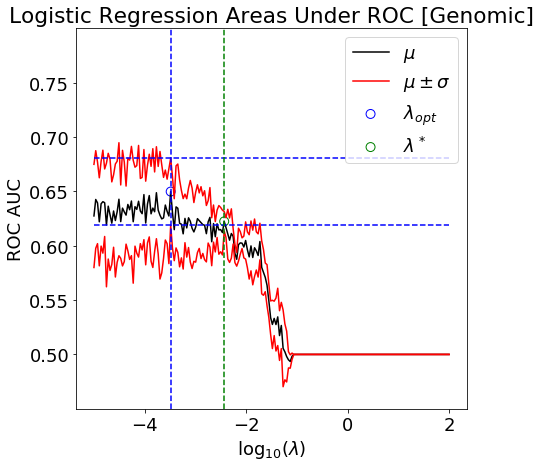}
  \caption{genomic}
  \label{fig:sub2}
\end{subfigure}
\caption{\textbf{Elastic Net Logistic regression hyperparameter tuning.}}
\label{fig:test}
\end{figure}

The parameter with the highest AUC is called $\lambda_{\text{opt}}$ and the chosen parameter is $\lambda^{*}$. With the chosen $\lambda^{*}$, the model achieved a test score of 64.8 on the clinical data and 61.9 on the genomic data. This is a significant improvement over our non-parametric approach, and the results are compared with other models in Tables 1 and 2.

\subsection{Support Vector Classifier}


We tried support vector classifiers with both linear and radial kernels, and the outcomes were similar in both cases. We tuned the parameter $C$, which restricted the number of data points that were allowed on the wrong side of their margin. Here, a larger $C$ produces a smaller margin and thus a more complex boundary. We might have expected more variance from overfitting due to large $C$s. At a certain point, we believe shrinking the margin did not change the classifications because the categorical or high-dimensional data was far from the separating hyperplane. As a result, the AUC scores plateaued. 

A small $C$ produces simpler models with a smoother boundary for the radial kernel. The simplest models achieved a minimal AUC of 50 as before. But the tuned linear model achieved results that were quite similar to the Logistic Regression model as seen in Tables 1 and 2. This is not suprising, given that logistic regression can be reduced to support vector machines \cite{zhou2015reduction}.

\begin{figure}[hb]
\centering
\begin{subfigure}{0.25\textwidth}
  \centering
  \includegraphics[width=\linewidth]{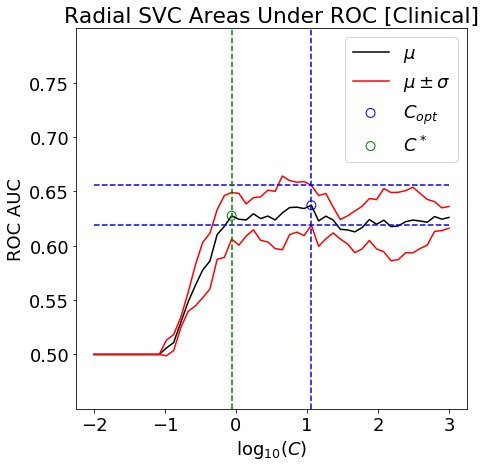}
  \caption{clinical}
  \label{fig:sub1}
\end{subfigure}%
\begin{subfigure}{0.25\textwidth}
  \centering
  \includegraphics[width=\linewidth]{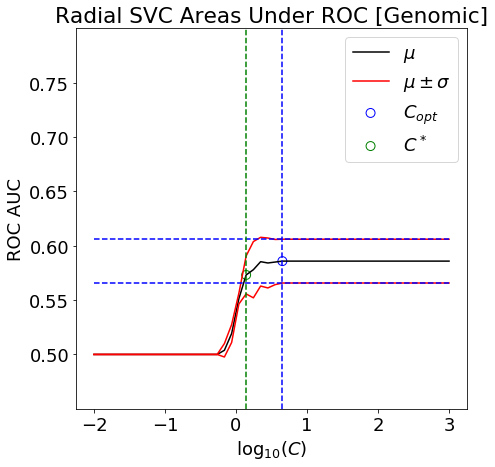}
  \caption{genomic}
  \label{fig:sub2}
\end{subfigure}
\caption{\textbf{Radial SVC hyperparameter tuning.}}
\label{fig:test}
\end{figure}

\subsection{Multi-layer Perceptron}

We tested the scikit-learn's built in MLP classifier to compare with our advanced implementation. We used 70 hidden layers, ReLU activation, and a consant default learning rate of 0.001. In the validation stage, we tuned the L2 regularization penality $\lambda$. However, the accuracy curves did not reveal any clear trends. If we had more time, we would also perform a grid search to optimize $\lambda$ as well as the number of hidden layers and the learning rates.

\subsection{Random Forests}

Lastly, we considered random forests. These are often considered strong candidates for learning algorithms because they are insensitive to features scales or transformations, and they are well-equipped to handle categorical data. For clinical data, we tuned the parameter $n$, which gave the number of trees in the forest. Individual trees have low bias and high variance, but when many trees are involved with bagging, the variance is reduced. On the clinical data, the algorithm performed about as well as logistic regression and SVMs. However, increasing the number of trees did not seem to improve scores. 

We did not run a full cross validation with the genomic data due to time constraints. But we see they achieve no better than random guessing on the test set. For an explaination, we reconsidered the algorithm. For each estimator, $m$ variables are selected at random from the total $p$ and the best split is picked from these. With $p \gg m$, we have a low probability of choosing a relevant feature with each split. To solve this problem, we would need to increase $m$.

\subsection{Summary of Baseline Algorithms}

\begin{table}[htbp!]
\centering
\caption{\textbf{Summary of Baseline Algorithms on Clinical Data.}}
\begin{tabular}{|l|r|r|} \hline
\textbf{Model}&\textbf{Val}&\textbf{Test} \\ \hline
K-nearest Neighbors & 59.7 & 55.8 \\ \hline
Logistic Regression & 66.5 & 60.1 \\ \hline
Linear SVC & 67.1 & \textbf{65.9} \\ \hline
Radial SVC & 67.2 & 64.8 \\ \hline
MLP Classifier & \textbf{68.3} & 65.4 \\ \hline 
Random Forest & 66.3 & 59.5 \\ \hline
\end{tabular}
\end{table}

\begin{table}[htbp!]
\centering
\caption{\textbf{Summary of Baseline Algorithms on Genomic Data.}}
\begin{tabular}{|l|r|r|} \hline
\textbf{Model}&\textbf{Val}&\textbf{Test} \\ \hline
K-nearest Neighbors & 50.7 & 52.1 \\ \hline
Logistic Regression & 60.1 & \textbf{61.9} \\ \hline
Linear SVC & 60.1 & 60.5 \\ \hline
Radial SVC & 60.5 & 56.8 \\ \hline
MLP Classifier & \textbf{62.6} & 57.8 \\ \hline 
Random Forest & N/A & 52.0 \\ \hline
\end{tabular}
\end{table}

In summary the models trained on the clinical data were not able to outperform those trained on the genomic data. In our stretch algorithms, we will try to improve on the encouraging results seen from the MLP and the generalized linear models. We will also try an implementation of XGBoost and a semi-supervised approach with logistic regression.

\section{Stretch Algorithms}
\subsection{Semi-Supervised Learning}
Traditional classifiers require labeled datasets to train, while it is often difficult to obtain such datasets. To advance machine learning in low-resource domains, semi-supervised learning has been widely applied. Semi-supervised learning is such a learning paradigm inspired by natural systems such as humans learn from both labeled and unlabeled data. 

In this project, we implemented two semi-supervised learning algorithms, namely self-training \cite{nigam2000analyzing} and co-training \cite{blum1998combining}. 

\subsubsection{Analysis of Self-Training from Our Perspective}

Self-training works in a clustering fashion, where a classifier trains itself on labeled dataset then teaches itself using its prediction of unlabeled dataset. Only the unlabeled samples with highest confidence are considered trustworthy to be new training instances. In this way, those unlabeled samples, which are closest to the labeled ones, or alternatively farthest to the decision boundary, are labeled with the same label as its neighbors. The workflow of self-training is formalized in Algorithm \ref{alg:self}.

\begin{algorithm}
\SetAlgoLined
\KwIn{Labeled dataset $\mathcal{L} = \left\{ {{{\bf{x}}_i},{{y}_i}} \right\}_{i = 1}^L$, unlabeled dataset ${\cal U} = \left\{ {{{\bf{x}}_j}} \right\}_{j = L + 1}^{L + U}$, confidence threshold $\alpha$}
\KwOut{Classifier $f \colon \mathbf{x} \to y$}
 fit $f$ to $\mathcal{L}$\;
 current prediction $\mathbf{\hat{Y}} \gets \left\{ f({{{\bf{x}}_j}}) \right\}_{j = L + 1}^{L + U}$\;
 previous prediction $\bf{\hat{Y}}^\prime \gets none$\;
 
 \While{${\bf{\hat Y}} \neq {{\bf{\hat Y}}^\prime }$}
 {
  ${{\cal U}^\prime } \leftarrow \left\{ {{{\bf{x}}_j},{\rm{confidence}}\,{\rm{of}}\,\left( {{{\bf{x}}_j},{{\hat y}_j}} \right) > \alpha } \right\}$ \;
  fit $f$ to $\left\{ {{\cal L},{{\cal U}^\prime }} \right\}$\;
  $\bf{\hat{Y}}^\prime \gets \bf{\hat{Y}}$\;
  $\mathbf{\hat{Y}} \gets \left\{ f({{{\bf{x}}_j}}) \right\}_{j = L + 1}^{L + U}$\;
 }
 \Return{$f$}\;
 \caption{Self-Training}
 \label{alg:self}
\end{algorithm}

The weakness in self-training is that the augmentation of data solely depends on the prediction of the unique weak learner being trained. Once the weak learner accidentally biases the decision boundary, its predictions are not reliable anymore. While in the 5th line of Algorithm \ref{alg:self}, those predictions are still added to the training set without more careful treatment, which might worsen the model. One can imagine self-training as a k-means clustering algorithm, with all labels in the training set removed temporarily.
The weak learner can be imagined as the clustering result produced with a random selected set of centroids. If the centroids are not well selected, the clusters will largely differ from the real distribution. If we labeled the clusters and use those as a KNN classifier, the prediction will mostly be biased too.

\subsubsection{Analysis of Co-Training from Our Perspective}

Instead of relying on a single classifier, co-training employs multiple classifiers trained on unequal views of the data. Under the assumption that each sub-view is sufficient for learning when we have enough labeled data, the predictions from classifiers can be augmented to create inexpensive labeled data. In decoding phase, classifiers $\left\{ {{f_k}:{{\bf{x}}_{\left[ {{d_k}} \right]}} \to y} \right\}$ are then ensembled to make a prediction using a voting strategy. Co-Training is illustrated in Algorithm \ref{alg:co}.

\begin{algorithm}
\SetAlgoLined
\KwIn{Labeled dataset $\mathcal{L} = \left\{ {{{\bf{x}}^{\left( i \right)}},{y^{\left( i \right)}}} \right\}_{i = 1}^L$, unlabeled dataset ${\cal U} = \left\{ {{{\bf{x}}_j}} \right\}_{j = L + 1}^{L + U}$, number of classifiers $n$, sub-feature set $\left\{ {{d_1}, \ldots ,{d_n}} \right\}$, max iteration $t$}
\KwOut{Classifiers $\left\{ {{f_k}:{{\bf{x}}_{\left[ {{d_k}} \right]}} \to y} \right\}$}
training sets ${{\cal T}_{\left[ {{d_k}} \right]}} \gets {{\cal L}_{\left[ {{d_k}} \right]}}$\;
\For{$iter \gets 1$ \textbf{to} $t$}
 {
  \For{$k \gets 1$ \textbf{to} $n$}
  {
   fit $f_k$ to ${{\cal T}_{\left[ {{d_k}} \right]}}$\;
   predict ${\widehat {\bf{Y}}^{\left( k \right)}} \leftarrow {f_k}\left( {{{\cal U}_{\left[ {{d_k}} \right]}}} \right)$
  }
  add samples $\widehat {y}_j$ to $\mathcal{T}$ where $\widehat {y}_j^1 = \widehat {y}_j^2 =  \ldots  = \widehat {y}_j^k$
 }
 \caption{Co-Training}
 \label{alg:co}
\end{algorithm}

With multi-view of the same training data, classifiers are working on conditionally independent dataset. Since the assumption is that given the labels, the sub-views are conditionally independent to each other. If one classifier knows the label $y$, then its sub-view will not help to guess the other sub-views. This condition prevents the weak learners from compromising to each other. In this way, line 7 in Algorithm \ref{alg:co} ensures that more reliable new samples are added to the training set.

\subsubsection{Results}

Logistic regression is used as both the baseline model and the weak learner in semi-supervised learning algorithm.
We tune the hyper parameters on randomly selected 10\% of the whole dataset. 
For co-training, we use only two learners.
Specially, the sub-views of co-training are selected through all combinations of two sets of features. 
Due to the large amount of computation, we only experiment co-training on clinical data.
The results on clinical data and genomic data are shown in Table \ref{tbl:semi}. 

\begin{table}[htbp!]
\centering\small\resizebox{\columnwidth}{!}{
\begin{tabular}{l||c|c}
\multicolumn{1}{c|}{\bf Model} & \multicolumn{1}{c|}{\bf Clinical} & \multicolumn{1}{c}{\bf Genomic} \\
\hline\hline
  Baseline   &     71.43 &     \textbf{65.00} \\
  Self-Training &     71.43 &  63.57 \\
  Co-Training &     \textbf{72.14} &  - \\

\end{tabular}}
\caption{Results on testset (last 10\% of whole dataset).}
\label{tbl:semi}
\end{table}

\subsubsection{Analysis}

\begin{figure}
            \centering
            \includegraphics[width = .45\textwidth]{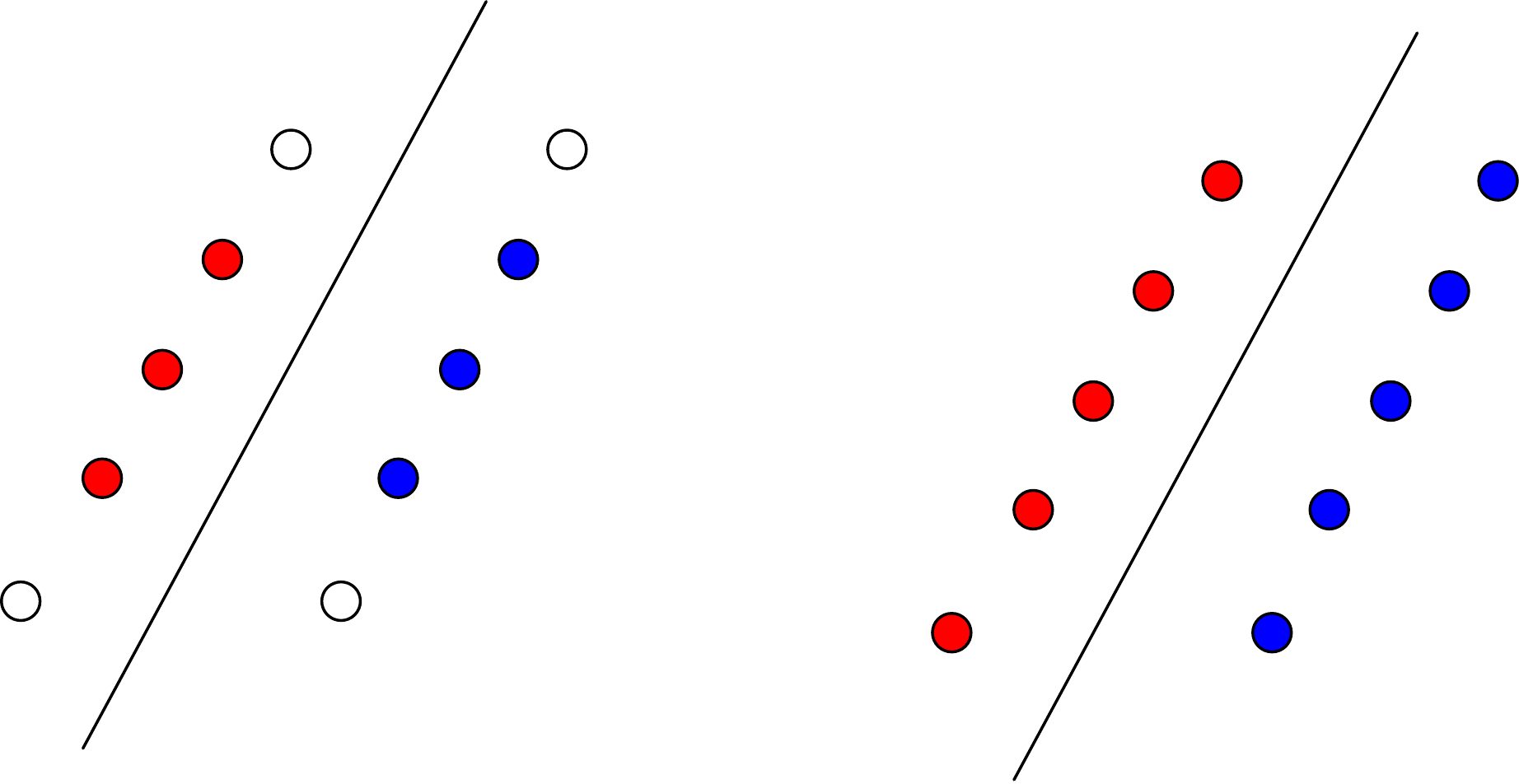}
            \caption{New samples does not affect decision boundary}
            \label{fig:self}
        \end{figure}

Not surprisingly, self-training shows similar or even worse performance against the baseline model. 
The reason might be that in low dimensional clinical data, unlabeled samples with high confidence are parallel to the decision boundary. 
Including those samples in training set creates no effect on the final decision boundary, as illustrated in Figure \ref{fig:self}.

While in high dimensional genomic data, chances for the samples to be parallel is much smaller.
As a result, those samples with high confidence tend to bias the decision boundary, leading to worse performance.

The co-training algorithm shows its advantage even with only 2 base learners. 
This result indicates the two assumptions for co-training hold. The first one assumes that the views are sufficient to predict the class well. The second one is that sub-views are independent given the class.

In our experiments, we find that one good sub-view which contains only 3 features, which are LYMPH\_NODES\_EXAMINED\_POSITIVE, CLAUDIN\_SUBTYPE and LATERALITY. Although lacking of experimental verification, we speculate that those three features are closely related to breast cancer.  
\subsection{L1-regularized Logistic Regression}

Given the relative success of baseline GLMs on both clinical and genomic data, we decided to implement our own logistic regression model using L1 regularization. The L1 penalty provides automatic feature selection by reducing irrelevant model weights to zero. The model is also interpretable, since weights can be ranked by feature importance. Lastly, the model is flexible and well-suited to a variety of learning problems.

As we can see in Tables 3 and 4, our model performed with similar success as the baseline scikit-learn algorithm. Our algorithm was implemented using iterative least squares whereas the scikit-learn algorithm was implemented using stochastic gradient descent. This may have caused some of the differences in results. 

\begin{table}[htbp!]
\centering\small\resizebox{\columnwidth}{!}{
\begin{tabular}{l||c|c}
\multicolumn{1}{c|}{\bf Model} & \multicolumn{1}{c|}{\bf Clinical} & \multicolumn{1}{c}{\bf Genomic} \\
\hline\hline
  Sklearn Baseline & 65.3 & 59.0 \\
  L1 Logistic Regression  & \textbf{69.2} & \textbf{61.6} \\
\end{tabular}}
\caption{AUC scores on val set}
\label{tbl:goamlptest}
\end{table}

\begin{table}[htbp!]
\centering\small\resizebox{\columnwidth}{!}{
\begin{tabular}{l||c|c}
\multicolumn{1}{c|}{\bf Model} & \multicolumn{1}{c|}{\bf Clinical} & \multicolumn{1}{c}{\bf Genomic} \\
\hline\hline
  Sklearn Baseline & \textbf{63.8} & \textbf{58.7} \\
  L1 Logistic Regression  & 60.5 & 52.9\\
\end{tabular}}
\caption{AUC scores on test set}
\label{tbl:goamlptest}
\end{table}

Both models both weighted the Lymph Node Count with highest priority. They also emphasized the "integrative clusters" which contained information about tumors. Our model emphasized age at diagnosis and the inferred menopausal state, whereas the baseline model emphasized NPI and breast surgery. In this, they differ.

Due to some errors in the code, our stretch algorithm may also be overfitting. We noticed that, despite the L1 penalty, none of the weights had been set to zero. The flat shape of the plots in Figures 6 and 7 also suggest that our model is not sensitive enough to the regularization parameter. We can see that this greatly affects the variance of the model, especially regarding the clinical data. For this reason, we should be wary of the reported clinical testing scores. 

\begin{figure}[ht!]
\centering
\begin{subfigure}{0.25\textwidth}
  \centering
  \includegraphics[width=\linewidth]{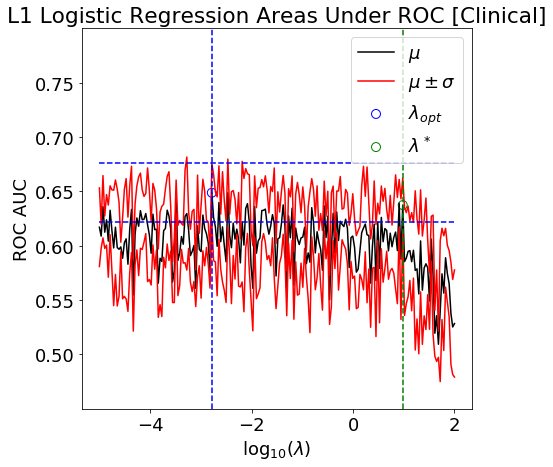}
  \caption{clinical}
  \label{fig:sub1}
\end{subfigure}%
\begin{subfigure}{0.25\textwidth}
  \centering
  \includegraphics[width=\linewidth]{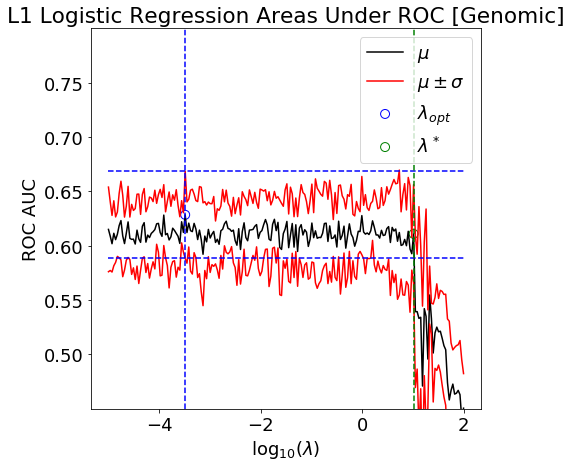}
  \caption{genomic}
  \label{fig:sub2}
\end{subfigure}
\caption{\textbf{Our L1-Regularized Logistic Regression model tuned using CV.}}
\label{fig:test}
\end{figure}

\begin{figure}[ht!]
\centering
\begin{subfigure}{0.25\textwidth}
  \centering
  \includegraphics[width=\linewidth]{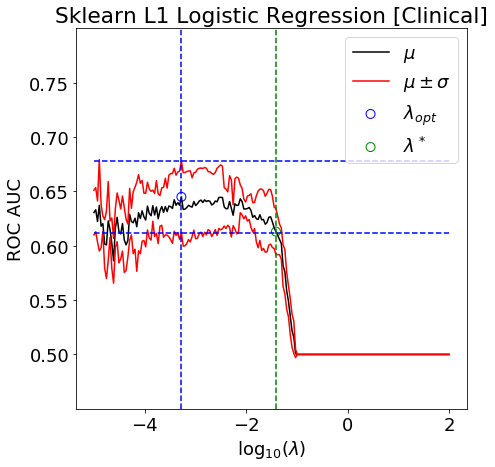}
  \caption{clinical}
  \label{fig:sub1}
\end{subfigure}%
\begin{subfigure}{0.25\textwidth}
  \centering
  \includegraphics[width=\linewidth]{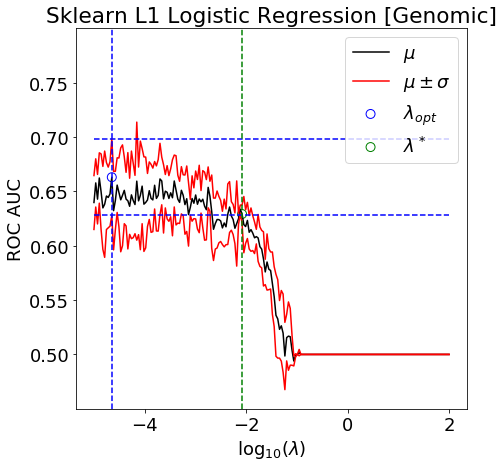}
  \caption{genomic}
  \label{fig:sub2}
\end{subfigure}
\caption{\textbf{Sklearn's L1-Regularized Logistic Regression model tuned using CV.}}
\label{fig:test}
\end{figure}

Overall, however, these experiments reaffirm one of the great advantages of logistic regression. The model is quite flexible and capable of accommodating very different types of data.

\subsection{Multilayer Perceptron with Grasshopper Optimization \cite{article} }
\subsubsection{Algorithm description}

The GOA is a novel optimizer that tries to inspire the social life of grasshopper insects in nature. The main behaviors of the grasshoppers are foraging, target pursuing, and team behaviors in either nymph or adulthood phases. In the larval level, they often exhibit short-length jumps with slow motions. In adulthood, they do long-range and swift movements to obtain food sources from farming areas. To simulate these facts, this optimization model was designed.

\subsubsection{Analysis of the Algorithm from Our Perspective}

For giving insights for how the algorithm really works, we analyze their algorithm step by step to see why this algorithm can converge and find the optimal value. Basically, every optimization's purpose is to find the best weights for a model. Here they encode the weights into a space vector which is the space position of a grasshopper, and this space position is constantly changed by some nature causes such as gravity, wind and the social communication among grasshoppers. However, these grasshoppers have their activity area, once they are over the boundary of the area, they will be justified back to their activity area. For selecting the best space vector in each iteration they compute the fitness function which is the object function of their optimization.

First we analyze how they compute the space position of a grasshopper in each iteration.  The new position of a grasshopper is attained according to its current location, the location of the specific target, and the situation of all population. The decreasing c factor assists GOA to gradually reduce the comfort zone. Hence, it can perform a smooth transition from exploration to exploitation of the fitness landscape. The repulsion forces can assist population for broad exploration of the fitness. These factors are important for adjusting how grasshoppers change their behaviours. 

Next we analyze how the selection algorithm works. We first initialize several grasshoppers and for each iteration, change their space position based on the algorithm and choose several the best performing grasshoppers for this iteration and put them into next iteration. After several iterations the weights will converge to a certain value. Why this works is because the weights will be changed by the factor of c and this c will be smaller and smaller to make the algorithm converge to a final convergence value. The problem of this procedure is that we cannot know which direction is good or bad even if you explore more in the first several iterations and explore less for last several iterations. For gradient based algorithm even if it cannot find the global optima, it always has good direction for local optimization, but for this algorithm, we cannot judge which direction is good or bad, since what they really control is the change extent of the weights and let the model itself figure out which weights are better than previous weights. 

\subsubsection{Result}
Table \ref{tbl:goamlpdev} shows the result on the dev set and Table \ref{tbl:goamlptest} shows the result one the test set. From the table, we know that on dev set the algorithm works better than original MLP, but in the test set, it drops a lot which makes it really inapplicable.

\begin{table}[htbp!]
\centering\small\resizebox{\columnwidth}{!}{
\begin{tabular}{l||c|c}
\multicolumn{1}{c|}{\bf Model} & \multicolumn{1}{c|}{\bf Clinical} & \multicolumn{1}{c}{\bf Genomic} \\
\hline\hline
  MLP Baseline   &     72.82 &     \textbf{68.45} \\
  Grasshopper MLP &     \textbf{74.11} &  \textbf{68.45} \\
\end{tabular}}
\caption{Results on val set}
\label{tbl:goamlpdev}
\end{table}
\begin{table}[htbp!]
\centering\small\resizebox{\columnwidth}{!}{
\begin{tabular}{l||c|c}
\multicolumn{1}{c|}{\bf Model} & \multicolumn{1}{c|}{\bf Clinical} & \multicolumn{1}{c}{\bf Genomic} \\
\hline\hline
  MLP Baseline   &     \textbf{70.70} &     \textbf{62.60} \\
  Grasshopper MLP &     68.03 &  56.74 \\
\end{tabular}}
\caption{Results on test set}
\label{tbl:goamlptest}
\end{table}
\subsubsection{Analysis of the Result}
The advantage of this algorithm is that it may avoid local solutions and find optimal results proportional to the number of iterations. The disadvantage is that it needs too many training iterations. In our experiment, For clinical data, we initialize 300 grasshoppers with 1000 iterations, after each iteration, the fitness always keeps going down, so we cannot accurately know where we should stop, so we just stop early to make it not so overfitting, but from the results, we can see that even if we stopped so early, it still was overfitting, so where we should stop is the big problem of this approach.

We found out four possible reasons of it's low accuracy on test set. 1.The data size is small and this complex model can be easily over fitting. 2.The time we stop is not appropriate since there is not a specific stop criteria. 3.The inappropriate chosen of fitness function since this is not the gradient based algorithm, in some extent, the fitness function can be any function that represent the difference between the prediction and label. 4.Some other implement details not mentioned in the paper, such as how to initialize the initial weights, so we just inicialize them randomly.

\subsubsection{Use GOA for Tuning Hyperparameter}

The basic method using GOA for hyperparameter tuning is to encode hyperparameters as a space vector of grasshopper and using the evaluaton metrics on dev set as the fitness function. However, you need to set different boundary of each hyperparameter which is now impossible for this algorithm, because this algorithm need to keep each dimension in the space vector have the same boundary to update the position of the grasshopper. we solved this problem by projecting the different domains into the same domain. For different hyperparameters, they have different domains. We project these domains into a domain with the same boundary using different weights and bias for each hyperparameter and when computing the value of the fitness function we project the same domain back to its original hyperparameters. 

The result of this algorithm on hyperparameter tuning is not as the expected though, it can be very high on dev set which is 77.50\% but in the test set it will drop to 65.53\% so using this kind of method to do hyperparameter tuning will generate huge bias between dev and test set, because apparently you can get the optimal hyperparameters on the dev set, but the best hyperparameters on the dev set are not the same as the best one on the test set.

\subsection{XGboost}
\subsubsection{Algorithm Description}

Xgboost is a popular gradient boosting library that has been winning many Kaggle contests in recent years. It is built under gradient boosting framework and optimized with its regularized objective function, approximate split finding algorithm, sparsity-aware split finding technique and elegant system design\cite{Tianqi2016Proceedings}.

\subsubsection{Analysis the Algorithm from Our Perspective}

\par\noindent \textbf{Parallel Computing:} Xgboost is built on gradient boosting framework which trains the model in a additive manner. However, additive training is a sequential algorithm, which is incompatible with parallel computing. To improve the efficency of the Xgboost, it parallelizes the algorithm in the tree building step. It can either parallelizes node building at each level of the tree or parallelizes split finding at each node. However, parallelizing in the node building at each level of the tree only performs as well as parallelizing split finding at each node when there is a huge data set.

\par\noindent \textbf{Regularized Learning Objective Function:} To improve Xgboost's performance, Xgboost adds regularized term to its loss function and combined them as a structure score function to measure how good a tree structure is. To ensure its scalability to various loss function, it introduce Taylor expansion of the second order of the loss function to the structure score function, making the value of the objective function only depends on the gradient and hessian statistics. This attribute also makes parallel computing and distribute computing easier.
\par\noindent \textbf{Approximate greedy Split Finding Algorithm:} Xgboost's high efficiency also benefited from its approximate split finding algorithm. Original exact greedy split finding simply enumerates over all possible splitting points greedily whereas approximate split finding proposes candidate splitting points according to percentiles of feature distribution, features are mapped into buckets split by these proposed points and then finds the best solution among proposals based on the aggregated statistics. However, sometimes algorithm may need to deal with weighted data set. Since there was no existing quantile sketch, Xgboost introduced a novel distributed weighted quantile sketch algorithm that supports merge and prune operations to deal with weighted data.

\par\noindent \textbf{Sparsity-aware Split Finding:} When dealing with real world problems, it is common for the input to be sparse. However, sparse data can be much different from normal data, since it may caused by artifacts of feature engineering such as one-hot encoding or caused by the presence of missing value or frequent zero entries. To make the algorithm be aware of the sparsity pattern in the data, Xgboost add a default direction for missing value in each tree node. The algorithm is designed to learn to find the best direction to handle missing values. It will first enumerate missing value to the right and compute the structure score of current node and then enumerate missing value to the left, getting the other structure score. Finally, it will choose the direction that earns better structure score.

\par\noindent \textbf{System Design:} Its system design contributes to its efficiency in low level. To reduce the time used in sorting data, Xgboost stores the data in in-memory units, named block. Data in each block is stored in the compressed column format, with each column sorted by the corresponding feature value.This input data layout only needs to be computed once before training, and can be reused in later iterations. It also apply cache-aware access and blocks for out-of-core computation to improve the speed of data fetching.

\subsubsection{Result}
Table \ref{tbl:Xgb dev set} shows the result on the validation set while Table \ref{tbl:Xgb test set} shows the result on the test set. 
\begin{table}[htbp!]
\centering\small\resizebox{\columnwidth}{!}{
\begin{tabular}{l||c|c}
\multicolumn{1}{c|}{\bf Model} & \multicolumn{1}{c|}{\bf Clinical} & \multicolumn{1}{c}{\bf Genomic} \\
\hline\hline
  Sklearn GBDT   &     68.62 &     60.53 \\
  Xgboost        &     \textbf{74.74} &     \textbf{73.38} \\
\end{tabular}}
\caption{AUC scores on val set}
\label{tbl:Xgb dev set}
\end{table}
\begin{table}[htbp!]
\centering\small\resizebox{\columnwidth}{!}{
\begin{tabular}{l||c|c}
\multicolumn{1}{c|}{\bf Model} & \multicolumn{1}{c|}{\bf Clinical} & \multicolumn{1}{c}{\bf Genomic} \\
\hline\hline
  Sklearn GBDT   &     60.57 &     53.10 \\
  Xgboost        &     \textbf{70.84} &     \textbf{67.12} \\
\end{tabular}}
\caption{AUC scores on test set}
\label{tbl:Xgb test set}
\end{table}

\begin{table}[htbp!]
\centering\small\resizebox{\columnwidth}{!}{
\begin{tabular}{l||c|c}
\multicolumn{1}{c|}{\bf Model} & \multicolumn{1}{c|}{\bf Clinical} & \multicolumn{1}{c}{\bf Genomic} \\
\hline\hline
  Sklearn GBDT   &     64.65 &     57.38 \\
  Xgboost        &     \textbf{73.36} &     \textbf{68.43} \\
\end{tabular}}
\caption{5 Folds CV avg AUC scores }
\label{tbl:Xgb 5 Folds CV}
\end{table}

Gradient boosting decision tree was selected as the baseline algorithm because both GBDT and Xgboost are based on gradient boosting framework. Xgboost earned obvious higher AUC scores on both clinical and genomic data set. Also, Xgboost yields smaller differences between the AUC scores on development set and test set. However, as least for our implementation, Xgboost takes much longer time to process genomic data. Last but not least, the parameters tuning was done solely on validation set as mentioned in chapter4.2.

\subsubsection{Analysis of the result}
Adding regularized term to its objective function helps Xgboost
avoid over-fitting and yields better results. The approximate greedy split finding and sparsity-aware algorithm improve Xgboost's efficiency and performance. Well designed data structure and algorithms enable Xgboost to perform parallel computing.The using of Taylor expansion in its objective function allow Xgboost to use different loss function as long as they have second order gradient. These advantages above make Xgboost a great tool of gradient boosting based machine learning tool.

However, we was not able to apply system design techniques, approximate greedy split finding algorithm to our implementation of Xgboost. Besides, the language we chose as implementation language was python which only supports multiprocessing and its efficiency can not compare with C++. What mentioned above make our implementation needs much more time than official Xgboost or Sklearn GBDT when dealing with huge data set.

\section{Conclusion}
In this project, we demonstrate a complete pipeline from preprocessing to analyzing the results. This project challenges the breast cancer survival prediction problem with both baseline models and stretch algorithms. 

We have seen that no model trained on the genomic data could be tuned to outperform our best clinical models. However, some approaches came closer than others. We see that a well-tuned ensemble method like XGBoost can perform competitively, despite the failure of our Random Forests in the baseline models. Xgboost also outperformed GBDT algorithm, since it emphasizes more on Regularization. Additionally, we found that our MLP classifiers were quite successful. These, like the ensemble methods, were capable of learning nonlinear decision boundaries in the complex and unrefined genomic dataset. One downside to MLP classifiers is that it may be more prone to overfitting. This could explain some of the larger differences in scores across the validation and testing sets. Another reason for this is that the validation and testing sets were not properly stratified to maintain class balances. The validation set contains 38.8\% Class 1 patients whereas the testing set contains 33.6\% Class 1 patients.

The success of the clinical models was largely due to the concentration of highly relevant features. Here, the generalized linear classifiers were much more successful. This may have been because the decision surfaces were simpler, or more likely, because the models were less prone to learning noise in the lower-dimensional space. In genomic data, the interrelations among different features make it much harder to achieve as successful result as in clinical data. Besides, the huge data set also make it more difficult for hyperparameter tuning.

Although not outstanding, the co-training approach shows promising compensation to the missing labels in clinical data. We also provide 3 features as a sufficient sub-view to predict breast cancer, which might be useful for medical researchers.

Two important future directions are dimensionality reduction or feature selection and grid-search hyperparameter tuning. The first is an important step for better handling any challenging high-dimensional data. The second is useful for training better models. Some of our best results came from the MLP classifier with GOA. We could push the limits of this success by performing a grid search on the many hyperparameters available. However, this would require significant time and computational resources. As always, these two factors of time and resources could make for potentially better results. Here we have demonstrated that the work may be worth the effort. Machine learning approaches can provide important insights for predicting clinical outcomes of breast cancer patients, and may be used in other medical settings for making key decisions.

\bibliographystyle{abbrv}
\bibliography{main}  
\end{document}